\documentclass[letterpaper, 10 pt, conference]{ieeeconf}  

\IEEEoverridecommandlockouts                              

\usepackage{todonotes}

\title{\LARGE \bf
Towards Mobile Multi-Task Manipulation in a Confined and Integrated Environment with Irregular Objects
}

\author{Zhao Han$^{\dagger}$, Jordan Allspaw, Gregory LeMasurier, Jenna Parrillo, Daniel Giger, \\S. Reza Ahmadzadeh and Holly A. Yanco$^{\dagger}$
	\thanks{All authors are with the Department of Computer Science, University of Massachusetts Lowell, 1 University Ave, Lowell, MA 01854, USA}
	\thanks{$^{\dagger}$Corresponding authors: {\tt\small\{zhan,holly\}@cs.uml.edu}}%
}

\begin{document}

\maketitle
\thispagestyle{empty}
\pagestyle{empty}

\begin{abstract}

The FetchIt! Mobile Manipulation Challenge, held at the IEEE International Conference on Robots and Automation (ICRA) in May 2019, offered an environment with complex and integrated task sets, irregular objects, confined space, and machining, introducing new challenges in the mobile manipulation domain. Here we describe our efforts to address these challenges by demonstrating the assembly of a kit of mechanical parts in a caddy. In addition to implementation details, we examine the issues in this task set extensively, and we discuss our software architecture in the hope of providing a base for other researchers. To evaluate performance and consistency, we conducted 20 full runs, then examined failure cases with possible solutions. We conclude by identifying future research directions to address the open challenges.

\end{abstract}

\section{Introduction}


Personal service robots have long been envisioned for work in people's homes. Although mobile robots have been deployed in open and structured environments such as warehouses and retail stores, realizing a truly useful household robot is still beyond the state of the art \cite{stuckler2016mobile}. 

Competitions can bring state of the art research into real-life scenarios. Robotic competitions have a history of spurring advancement of both research and implementation, as well as future personnel trained. One notable example is the 2005 DARPA Grand Challenge
where the competitors went on to make many contributions to research and industry (e.g., members the winning team, led by Stanford \cite{thrun2006stanley}, went on to head Google's self-driving car unit \cite{thrungoogle}).

We entered the 2019 FetchIt! mobile manipulation challenge \cite{fetchitws}, organized by Fetch Robotics and held at ICRA 2019, winning second place. The goal of the FetchIt! competition was for a Fetch mobile manipulator robot \cite{wise2016fetch} to navigate in a narrow work cell in order to assemble a caddy kit from complex mechanical parts such as gearbox pieces and screws lying on tables and in containers, requiring the  machining a large gear, then to transport the completed kit for inspection. The competition was designed to promote the state of the art mobile manipulation use in manufacturing and related applications \cite{mediaws}. Rather than focus on a specific task (e.g., pick and place in the Amazon Robotics Challenge), the competition task design encompasses the full range of activities commonly found in a manufacturing environment. 
Despite the industry-oriented design, the underlying challenges are also applicable to a household environment.

The contributions of this work are the implementation and software architecture details, an extensive analysis of the challenges posed while achieving such an integrated mobile multi-task manipulation, and suggestions for future research. 

\begin{figure}[t]
\centering
\includegraphics[width=0.9\columnwidth]{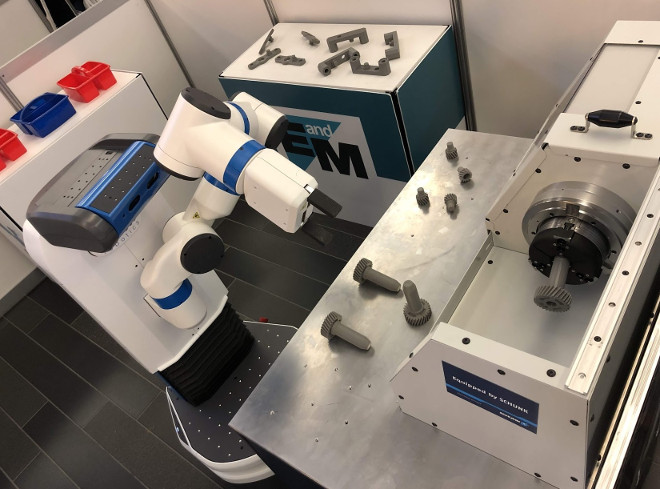} 
\caption{The competition arena, where the Fetch robot autonomously drives close to the gear station, grasps a large gear, inserts it into the lathe chuck, and retreats its arm to prepare for navigation.}
\label{fig:intro} 
\end{figure}


\section{Related Work}






Detecting and picking objects, also known as bin picking, has been investigated extensively with a fixed manipulator arm grasping objects detected by a 3D vision sensor (e.g., ~\cite{ikeuchi1983picking, rahardja1996vision, liu2012fast, schwarz2017nimbro}). To enable robots to perform similar tasks in unstructured human environments, we must extend these methods and frameworks for mobile manipulation. 

Several groups have developed mobile manipulation frameworks for finding, collecting, and delivering objects in kitchens (e.g., ~\cite{srinivasa2010herb, vahrenkamp2012simultaneous, beetz2011robotic}). Chitta et al. studied pick-and-place of ordered and well-separated objects using a mobile manipulator~\cite{chitta2012perception}; tasks included beverage fetching and object transportation. Although in most cases objects are ordered and well-separated, Nieuwenhuisen et al. studied bin picking with a mobile robot on an unordered set of objects~\cite{nieuwenhuisen2013mobile}. St{\"u}ckle et al. extended this framework using a mobile manipulator in a spacious room for the Robot@Home competition \cite{stuckler2016mobile} to grasp sausages from a barbecue using tongs, open bottles, and water plants. Pavlichenko et al. developed a mobile manipulation framework for part kitting in a car manufacturing environment~\cite{pavlichenko2018kittingbot}, focusing on part detection and arm trajectory optimization to increase the safety of humans in human-robot collaboration tasks. 

Most of these frameworks focus on specific aspects of a complex task and neglect other aspects. However, for robots to be effective and functional, a comprehensive framework must be designed. In recent years, several robot competitions have been designed to serve as benchmarks and encourage research (e.g.,~\cite{gerndt2015humanoid, guizzo2015hard, stuckler2016nimbro}). Our framework is designed for the FetchIt! Competition ~\cite{fetchitws} which requires navigation in a confined environment and interaction with an actual SCHUNK machine that has a confined operation space.

\section{FetchIt! Mobile Manipulation Challenge}

\begin{figure}[t]
\centering
\includegraphics[width=0.9\columnwidth]{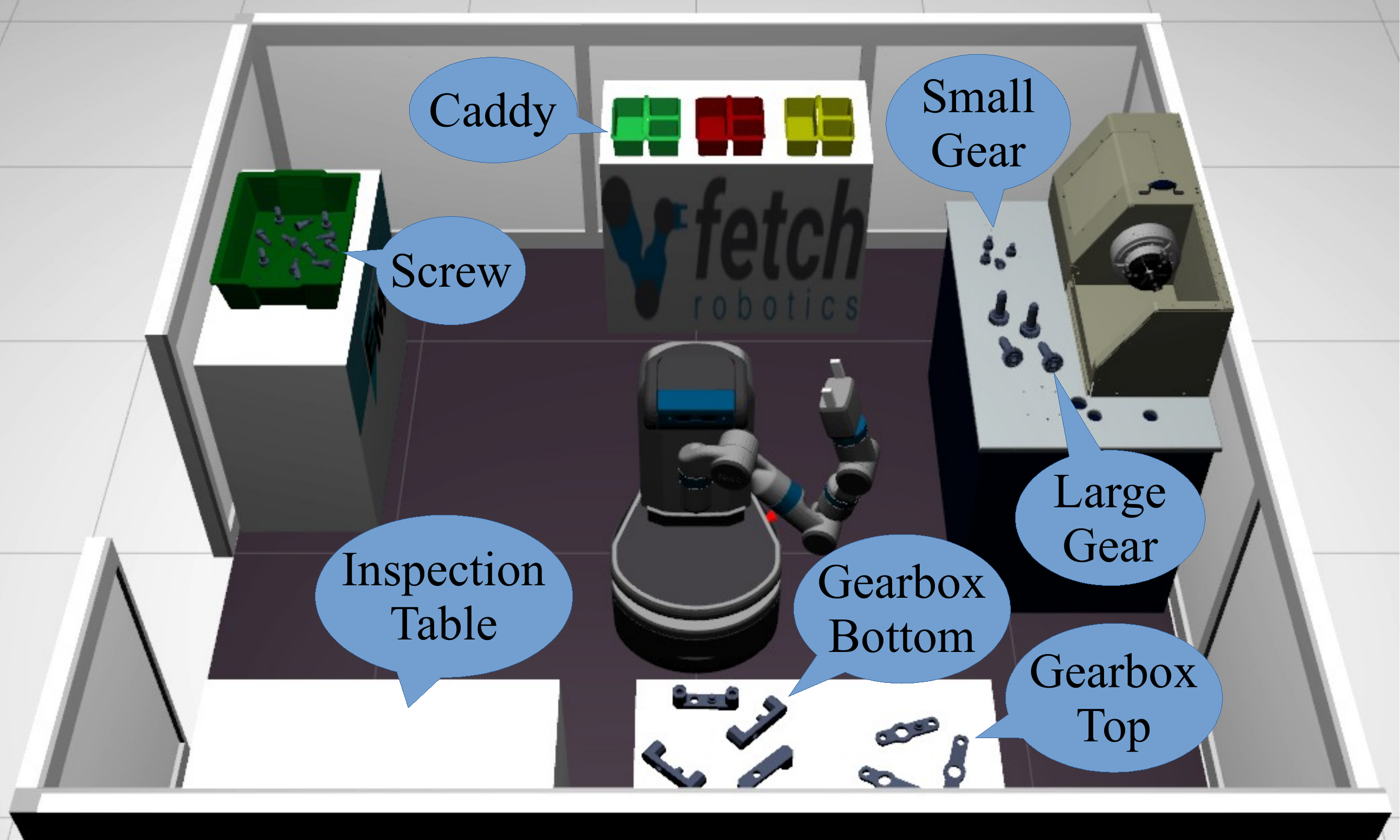} 
\caption{The FetchIt! Mobile Manipulation Challenge environment in simulation. The main goal is to place a specified set of parts into the correct sections of the caddy, then to transport the caddy to the inspection table.
}
\label{fig:sim_arena}
\end{figure}

The goal of the FetchIt! Challenge is to gather a set of objects into a caddy and deliver it to a designated area. The competition environment (Fig. \ref{fig:sim_arena}) is a $3.05 \times 3.05 m^2$ arena with walls. Except for the SCHUNK machine table, all other tables are of $0.785m$ height with a $0.46 \times 0.92 m^2$ tabletop. The SCHUNK machine table is $0.795m$ high with a $0.7 \times 1.1 m^2$ tabletop. The robot can interact with six types of objects, including small gears (4), large gears (4), gearbox tops (4), gearbox bottoms (4), bolts (10), and caddies (3).



\begin{figure}[t]
\centering
\includegraphics[width=0.8\columnwidth]{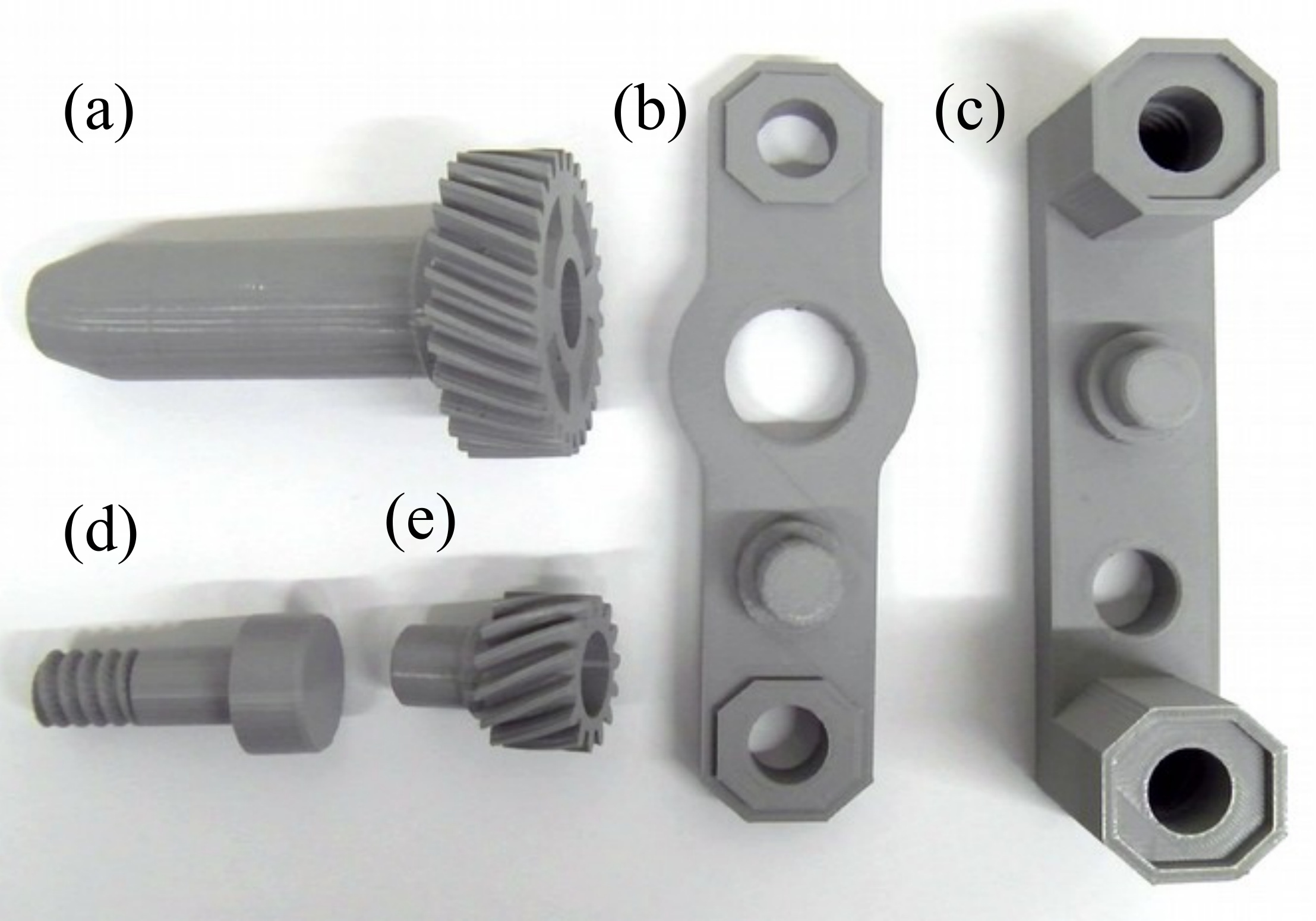} 
\caption{Parts to be collected: (a) Large gear (b) Gearbox top (c) Gearbox bottom (d) Screw (e) Small gear.}
\label{fig:parts}
\end{figure}




Fetch is a mobile robot with a single chest-mounted 7DOF arm and a head-mounted RGBD camera. Its base has a $30cm$ radius. The robot can adjust its torso vertically from a minimum height of $1.1m$ to a maximum of $1.5m$. The robot's maximum reachability is $114cm$ but it will reduce to $83cm$ when its wrist or gripper is pointing down. Because the arm is mounted on the chest, if counting the base radius, the maximum reachability is further reduced to around $53cm$. This is important because the robot has to navigate to a close vicinity of tables in order to reach target objects. The robot is also equipped with a Primesense Carmine short-range RGBD sensor with a $0.35$ to $1.4m$ range.




The goal of the competition was to score as many points as possible during a 45-minute run. The delivery of a complete caddy, all of the appropriate pieces in their correct compartment, received 7 points; no points were earned for incomplete caddies. A bonus point is awarded for delivering the caddy into a designated area underneath a camera on the inspection table (Fig. \ref{fig:sim_arena}). One point was deducted if a piece was placed in the wrong compartment, as well as for any extra pieces inside. One point was deducted for each item dropped to the floor, and for each non-trivial collision, either during navigation or manipulation.

\section{UML-HRI Mobile Manipulation Framework} 

In this section, we describe our developed framework and describe our implementation details for different modules including perception, object manipulation, and navigation.

\subsection{System Overview}

We used the Robot Operating System (ROS) framework~\cite{quigley2009ros} for the development of all the modules. For perception, we used the Point Cloud Library (PCL)~\cite{rusu2011point} to process the point cloud acquired from the head camera. For manipulation, we chose the MoveIt~\cite{chitta2012moveit} motion planning library for handling arm trajectories after specifying waypoints. For obstacle avoidance, we used the built-in octomap plugin to represent the scene as an octo-tree. To allow minor collisions with the mechanical parts for grasping, we customized the point cloud input of the octomap to build the scene from multiple views and to crop small areas. We also specify in-hand objects as collision objects, which is used for the large gear tasks. For navigation, we employed the ROS navigation stack~\cite{marder2010office} which uses a voxel grid to represent a known map and works based on the adaptive Monte Carlo localization~\cite{fox2002kld}, the A* based global planner~\cite{konolige2000gradient} and the Dynamic Window Approach local planner~\cite{fox1997dynamic}. For error handling, 5 retries are implemented during  motion planning and execution failures in addition to detection and navigation.  


\subsection{Perception}

As a common step for all of our perception procedures, a planar segmentation is performed and the resulting point cloud over the table is passed to the Euclidean clustering algorithm parameterized by the space between objects.

\subsubsection{Caddy Detection}

The robot is faced with three caddies (Fig. \ref{fig:sim_arena}), each with two small and one large compartments designated for specific parts. The robot uses a single caddy during each run. Each caddy can be placed on the table in any orientation, which is important for the caddy delivery phase and also specifies the location of the three compartments. 

Given the CAD file of the caddy, we initially tried to cluster the caddies and match the point cloud to the CAD model. To achieve this goal, we tested three different algorithms that were shown to be successful in detecting similar objects (e.g., animal sculptures). These algorithms consist of correspondence grouping \cite{chen20073d,tombari2010object}, hypothesis verification \cite{aldoma2012global}, hypothesis rejection \cite{buch2013pose}, none of which could detect the orientation of the caddy correctly and consistently. We did not attempt the convolutional neural network method \cite{krizhevsky2012imagenet} as data collection is difficult for the unknown competition arena and the learned model may not transfer to the competition arena environment.

Instead, we use a height heuristic to crop the point cloud of the caddy handles, cluster them using Euclidean cluster extraction, determine the $y$ axis using PCA, and determine the divider by checking points at both sides of the caddy.

\begin{figure}[t]
\centering
\newcommand\implheight{2.8cm}
\includegraphics[height=\implheight]{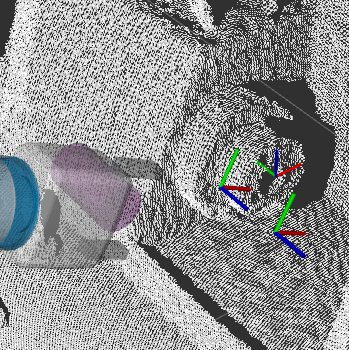}
\includegraphics[height=\implheight]{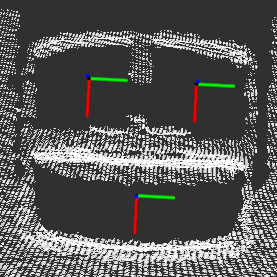}
\includegraphics[height=\implheight]{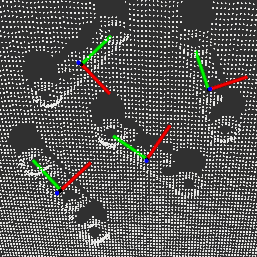}
\caption{Images from RViz showing three detection outputs. Left: In-hand large gear obstacle, two waypoints and extracted chuck pose. Middle: Poses of the three caddy compartments. Right: poses of four gearbox bottom parts.}
\label{fig:perception}
\end{figure}

\subsubsection{Mechanical Part Detection} 

Given the parts that the robot must manipulate (Fig.~\ref{fig:parts}), an intuitive method for clustering objects is to determine the part type by checking the dimensions of the cluster (i.e., width, length, and height). However, this method does not work for all cases: e.g., when the gearbox top or bottom pieces lay on their side, the simple dimension check is insufficient. An additional step should be taken if a piece was particularly difficult to distinguish, which can happen in several cases; for example, sometimes part of an object is removed during tabletop segmentation. In such cases, our system made a decision by looking at neighboring pieces, assuming that same type objects are placed close to each other. Ambiguous pieces were only selected if no ideal candidates could be found. To account for RGBD sensor noise, several frames of point clouds are captured and compared. If no candidates were detected within several frames, the robot would report the failure and try again.



\subsubsection{Chuck Detection}

The SCHUNK machine has a chuck, which the robot must be able to detect in order to operate the machine. Our detection method relies on the concept of horizontal and vertical planes. We first detect the bottom plane of the machine and the vertical plane to which the chuck is attached. Then the chuck is segmented and we find the rightmost plane of the segment using the dimension heuristic. The hole pose is calculated by averaging the four extreme points on the plane. The left image of Fig.~\ref{fig:perception} shows an example of the extracted chuck pose. 

\subsubsection{Machine Door Handle Detection}
For closing and opening the door on the SCHUNK machine, the algorithm first finds the largest cluster on the table and then uses the top point of the largest cluster to infer where the handle is using the handle height.


\subsection{Manipulation}
As shown in Fig.~\ref{fig:parts}, there are five types of parts. The robot needs to retrieve two screws from the bin on the left table in Fig. \ref{fig:sim_arena}; screws were guaranteed to not touch. The robot also needs to retrieve each of the gearbox top and bottom pieces on the bottom-middle table in Fig. \ref{fig:sim_arena}. Each type of gearbox piece is assumed to be clustered together. The right table in Fig. \ref{fig:sim_arena} contains four large and four small gears. Like the gearbox pieces, the gears are guaranteed to be clustered near each other; however, each cluster could be anywhere on the table. The gears could be lying down in any orientation or standing straight up. The parts would be replenished before they ran out, so there are always parts on the table.

\subsubsection{Picking Mechanical Parts} 
Once a top-down grasp pose (i.e., position and orientation) was determined from the object pose, a set of waypoints was created including one directly above the object and another directly on the grasp point. Before motion planning, the robot first looks around (in different directions), scanning the walls, tables, and other potential obstacles, and adds them to its current octomap. To allow collision with the object to be grasped but not the whole collision scene, the end-effector would pause directly above the object, then remove a partial or entire octomap based on how open it was above the grasp point. Then, the robot would raise its arm above the table, restore the excluded octomap, and calculate the trajectories. The main reason to perform this process is because the planner could see the grasped object as a collision and refuse to calculate a solution. Another reason to include intermediate waypoints between the manipulator start pose and the grasp point was to prevent the robot from performing unexpected rotations, such as rotating the elbow joint (e.g. sometimes by 179 degrees), while hovering over the table. After grasping an object, the gripper state was checked to verify that the object was grasped. If the gripper was closed entirely, meaning no object was grasped, the robot would attempt to grasp another object.


\subsubsection{Placing Parts into the Caddy}

After grasping parts, the robot needed to navigate to the caddy station to drop each type of object into different compartments. As long as the navigation to the target location is successful and the robot can detect the caddy, it will plan a trajectory to move the arm to a point above the right compartment and will open its gripper to release the object.

\subsubsection{Delivering the Caddy}
Once kitting was complete, the caddy needed to be picked up and transported to the inspection table on a marked area. Note that this marked area was not shown in simulation or before the competition. Additionally, during the competition, an aluminum inspection camera pole was added to the left of the inspection table for verification of the results. Because the marked area is either on the left or right side, we detect the corner of the inspection table by checking the front left or right points in the tabletop plane.


\subsubsection{Machining Gears}


To machine a large gear using the SCHUNK machine, the robot needs to insert the gear into the lathe chuck, achieved by specifying waypoints. Because the diameter of the hole is only $1cm$ larger than the shaft and due to inevitable grasping errors due to sensor noise and occlusion, we used the joint effort values to allow the robot to retry within a $4cm$ range of the detected hole. To avoid the robot pushing too much into the chuck when blocked, its arm only moves $2cm$ every time. After the gear is placed in the chuck hole, the robot sends a ROS service call to lock the chuck. Then the robot releases the gear and removes the arm from the machine via the same set of waypoints. Then the robot can optionally close the machine safety door by pushing the handle on top of the machine to the right.

The machining process is done in 5 minutes; the robot can check the status by sending a ROS service call. The robot can open the door if needed. The process of grasping the gear is the same as inserting it except that the end pose is shifted leftwards. A ROS service call releases the gear.


\subsection{Navigation}

We built our navigation module based on the existing ROS navigation stack~\cite{marder2010office} originally designed for an indoor office environment. The existing navigation stack had to be improved because it did not allow the robot to move close to a table (i.e., less than $20cm$) due to a rough base model~\cite{ciocarlie2012mobile}. Without improvements, the robot will get stuck and keep rotating for localization even if the inflation radius is reduced. Although the exisiting ROS navigation stack allows us to manually move the robot closer to the table, the competition required the robot to be fully autonomous. We addressed this issue by implementing autonomous manual base movements, which allows the robot to move towards the table after reaching an approximate goal, then moves away after finishing the manipulation task. This method is achieved by averaging a range of base lidar  scans. We also implemented table alignment to account for navigation rotation error using a scan range. In addition, to be able to adapt to changes in the environment when tables are rearranged, our system uses map poses specified by distances to nearby walls and tables.


\section{Software Architecture}

\begin{figure}[t]
\centering
\includegraphics[width=0.95\columnwidth]{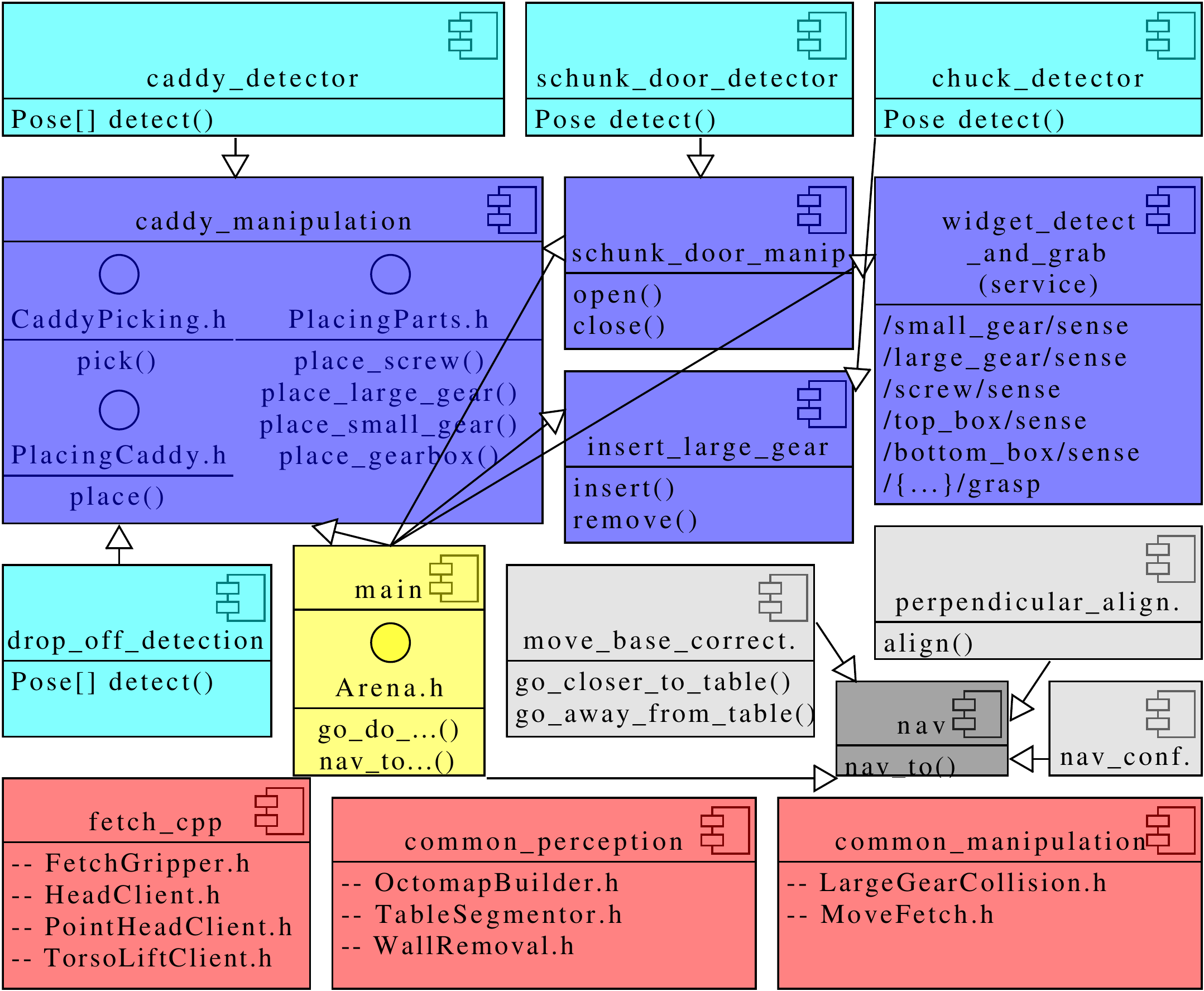} 
\caption{The Unified Modeling Language (UML) diagram of our architecture. Yellow, blue, cyan, gray and red blocks indicate the main, manipulation, perception, navigation and common code components respectively.}
\label{fig:arch_diagram}
\end{figure}

Our software architecture is largely object-oriented and component-based. A component is a catkin C++ library, where catkin~\cite{foote2013groovy} is a CMake based build system. Like any other C++ libraries, users can simply include the header files and use a component by initializing an object and calling the corresponding API methods.

Fig. \ref{fig:arch_diagram} illustrates our component-based architecture. Each component is a block that shows the component name and its interfaces. As shown in yellow, the main component comprises four manipulation components for the tasks and one navigation component. The manipulation components are separated from perception, and the navigation in dark gray is supported by a few other components. Common code components are shown in red with only interface files.

\section{Experimentation}

\begin{figure*}[th]
\centering
\newcommand\testcourseheight{2cm}
\includegraphics[height=\testcourseheight]{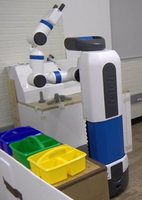}%
\thinspace%
\includegraphics[height=\testcourseheight]{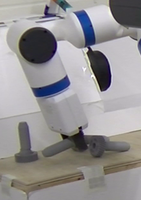}%
\thinspace%
\includegraphics[height=\testcourseheight]{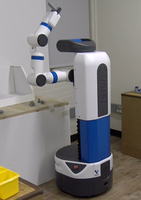}%
\thinspace%
\includegraphics[height=\testcourseheight]{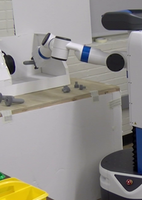}%
\thinspace%
\includegraphics[height=\testcourseheight]{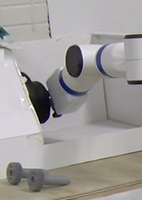}%
\thinspace%
\includegraphics[height=\testcourseheight]{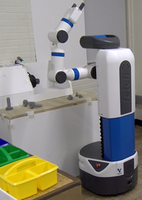}%
\thinspace%
\includegraphics[height=\testcourseheight]{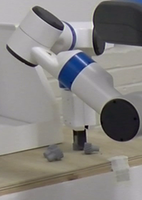}%
\thinspace%
\includegraphics[height=\testcourseheight]{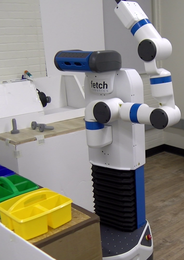}%
\thinspace%
\includegraphics[height=\testcourseheight]{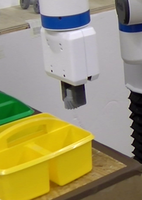}%
\thinspace%
\includegraphics[height=\testcourseheight]{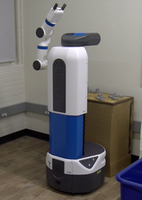}%
\thinspace%
\includegraphics[height=\testcourseheight]{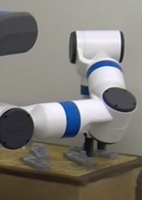}%
\thinspace%
\includegraphics[height=\testcourseheight]{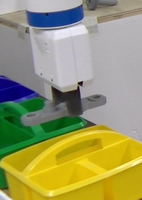}%
\\\vspace{.16667em}%
\includegraphics[height=\testcourseheight]{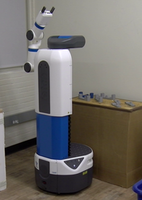}%
\thinspace%
\includegraphics[height=\testcourseheight]{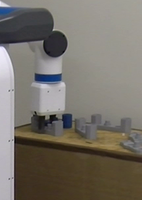}%
\thinspace%
\includegraphics[height=\testcourseheight]{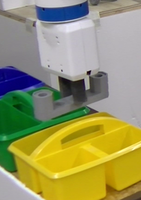}%
\thinspace%
\includegraphics[height=\testcourseheight]{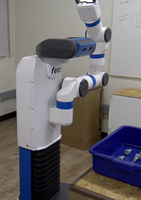}%
\thinspace%
\includegraphics[height=\testcourseheight]{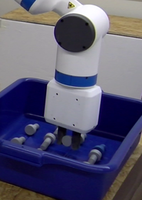}%
\thinspace%
\includegraphics[height=\testcourseheight]{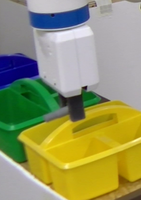}%
\thinspace%
\includegraphics[height=\testcourseheight]{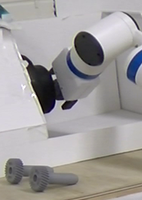}%
\thinspace%
\includegraphics[height=\testcourseheight]{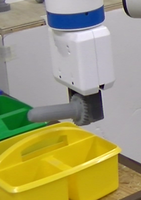}%
\thinspace%
\includegraphics[height=\testcourseheight]{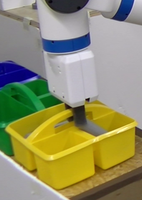}%
\thinspace%
\includegraphics[height=\testcourseheight]{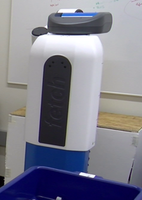}%
\thinspace%
\includegraphics[height=\testcourseheight]{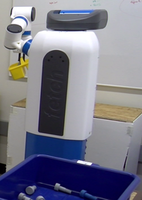}%
\thinspace%
\includegraphics[height=\testcourseheight]{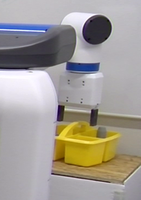}%
\caption{Snapshots of a successful kit delivery in our test course from the accompanying video \cite{video}.
}
\label{fig:sucess_run}
\end{figure*}

\begin{figure}[th]
\centering
\includegraphics{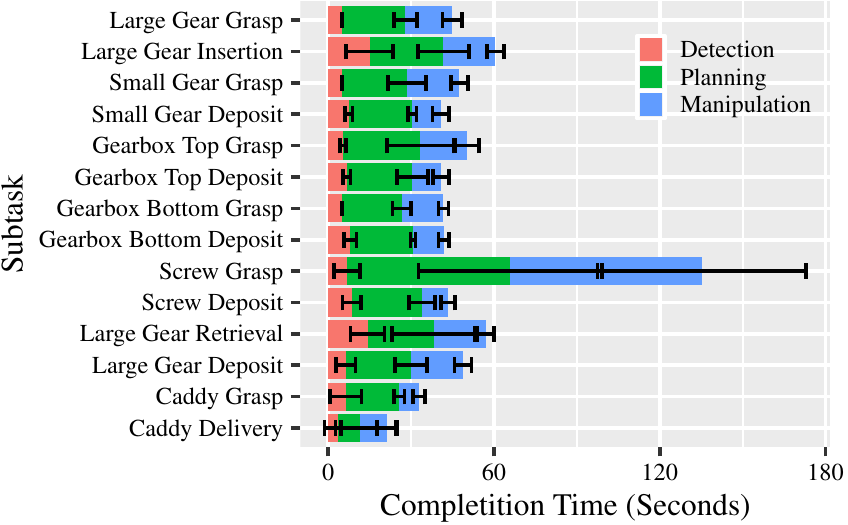} 
\caption{Time needed for each task with standard deviation error bars.}
\label{fig:bar_chart}
\end{figure}

After the competition, we continued to improve the code base, then evaluated the performance. We conducted 20 non-stop full runs in our test course and collected data from metrics such as task completion time, perception time, motion planning time, manipulation time, and task failure rate. We also recorded recoverable and non-recoverable failures. Fig. \ref{fig:sucess_run} shows images from a successful run.

We were able to fully assemble a kit per the competition rules in 13 of the 20 recorded runs.
Fig. \ref{fig:bar_chart} shows the time to complete each sub task, broken down into time spent on detection of objects, planning and manipulation. Planning includes time spent building the collision avoidance map and computing the inverse kinematics solution.

The challenges posed in this competition led to 7 failed runs where the robot failed to assemble and deliver the caddy: in 2 of these 7 runs, the chuck was not detected due to navigation error; in another run, the robot was unexpectedly backing up while it was stuck; in another, the small gear was slightly placed outside of the small compartment due to point cloud noise; in another, the large gear was grasped at an awkward angle due to occlusion; and in the remaining two failed runs, the in-hand large gear collided with the chuck due to occluded chuck back, which was considered free space.
Most of those issues could be solved with adjustments to the various detection algorithms to increase accuracy. The navigation error is very uncommon, as there are recovery behaviors in place, but one fix would have been to force the robot to abandon its current mission and attempt to navigate to the center of the arena, a relatively safe known area, then retry the task in the event of a navigation failure.

While there were minor failures in the 13 successful runs due to noisy RGBD sensor values and collisions at a specific execution waypoint (e.g., grasping a screw at the edge of the bin), they were all recovered within 5 retries. 
As seen in Fig. \ref{fig:bar_chart}, the standard deviation for some tasks was quite high because of the retries. For example, the fastest time to plan a grasp for a screw was 29 seconds; however, since the robot could try for several different screws before preceding, the slowest screw planning time was 150 seconds. While re-detecting objects is not time consuming, continuously re-planning and executing movement commands to reach an object requires more time. While the screw grasping task proved the biggest offender because the screws might be near the edge of the bin, several other tasks also relied on several retries during many of the runs. Another example was the planning step of the caddy deposit, which often took approximately one second; however, in several instances took twenty or forty seconds before the retrying step timed out and it returned back to the detection step. While ``retry until you succeed'' seemed very successful and mostly reliable where it was implemented, one improvement would be to better predict when an action is likely to fail before doing the work of planning which would have saved a significant amount of time.

\section{Challenges}


\begin{figure}[th]
\centering
\newcommand\implheight{2.8cm}
\includegraphics[height=\implheight]{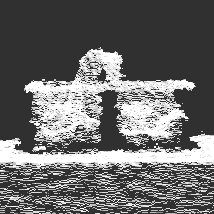}
\includegraphics[height=\implheight]{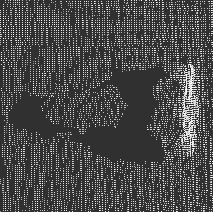}
\includegraphics[height=\implheight]{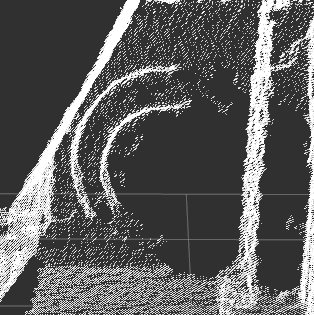}
\caption{A set of perception challenges. See Section~\ref{sub:perception_challenge} for more details. }
\label{fig:perception_challenges}
\end{figure}

\subsection{Perception Challenges}
\label{sub:perception_challenge}


\subsubsection{Caddy Detection} As illustrated in Fig. \ref{fig:parts}, the objects are complex and irregular. The caddy is a concave object with a curved handle and three compartments. As seen in Fig. \ref{fig:perception}, it is practically symmetric but one side has a divider that can be easily omitted due to occlusion, which is the case when the handle is horizontal and the divider might be behind the handle. Also, the divider becomes hard to recognize when a tall large gear is placed in it upright (Fig. \ref{fig:perception_challenges} left).

\subsubsection{Parts Detection} Since the Fetch robot does not have a camera on its hand to assist with visual servoing, we rely entirely on detecting the grasp pose of the object correctly. Due to the different dimensions of the gears, distinguishing them was not a significant challenge. However, because the large gearhead is bigger than its shaft, it could cause occlusion when the shaft was not facing the robot (see Fig.~\ref{fig:perception_challenges} middle). The occlusion causes the large gear to be seen as two separate objects in the point cloud. 

\subsubsection{Machine Chuck Perception and Manipulation} The circular shape of the chuck makes perception easier but there is a reachability issue: if the robot stops farther from the machine in order to see the hole, its arm cannot reach the chuck to insert the large gear. On the other hand, if the robot stops closer to the chuck for better reachability, it cannot see and detect the chuck (Fig.~\ref{fig:perception_challenges} right). When the chuck is visible to the robot, the distance between the robot and the chuck is around $90cm$, which is slightly longer than the $83cm$ arm length when the wrist is pointing downwards.

\subsection{Manipulation Challenges}

\subsubsection{Manipulation of Screws}
The screws were placed inside the bin at varying rotations. While they were guaranteed not to be touching, they could still be close together so careful grasping was required. In addition, the screw needed to be grasped without bumping the elbow on the bin, which, while accounted for with the octomap, led to a large number of failed inverse kinematic solutions and retry attempts.

\subsubsection{Manipulation of Large Gears}
Large gears could be on the table in two configurations: lying down on the table or standing up on the gear wheel. Grasping a gear in the second configuration is a challenge due to the robot's bulky palm and short fingers that can cause collision with the table. We chose to grasp the gears in the first configuration. 

Because the large gear had to be inserted straight into the machine chuck, a very accurate straight grasp angle on the wheel is required. When the arm straightens out after lifting, the gear would be perpendicular to the table and line up with the chuck hole. This proved to be a difficult approach, as often some or all of the gear would be obscured, leading to a slightly incorrect angle. However, this error was largely counteracted by using joint effort feedback. Because the large gear might be oriented incorrectly during grasping, it requires retries after failed insertions. 

\subsubsection{Obstacle Avoidance}

Operating in a confined space like the SCHUNK machine compartment poses challenges for obstacle avoidance. The back of the chuck is not captured by the head camera and thus deemed collision-free. The chuck plane is not fully visible because of the point cloud being sparse and noisy. The large gear cannot be accurately modelled in the collision scene. These reasons lead to in-hand large gear collision with the chuck, especially at back and the edges. Because the chuck back takes a decent amount of space in the free space in the small compartment, the chance of minor collisions is high when reaching multiple waypoints for joint effort checking.


\subsection{Navigation Challenges} 
\label{sub:nav_challenge}

\begin{figure}[t]
\centering
\newcommand\implheight{2.8cm}
\includegraphics[height=\implheight]{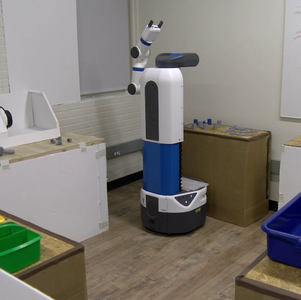}
\includegraphics[height=\implheight]{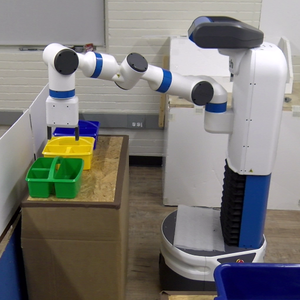}
\includegraphics[height=\implheight]{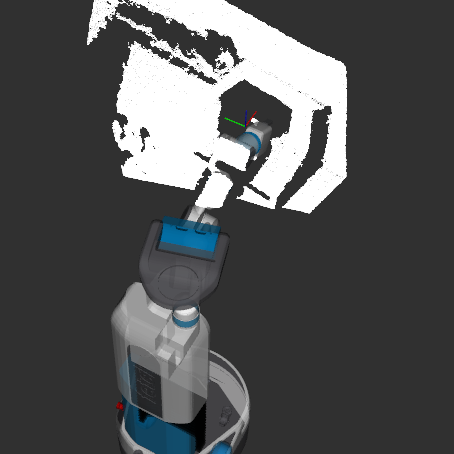}
\caption{A set of navigation challenges. See section~\ref{sub:nav_challenge} for more details.}
\label{fig:nav_manip_challenges}
\end{figure}

The robot has to navigate in a confined space which is roughly nine times bigger than the base of the robot ($0.6m$ in diameter). The large base limits the robot arm's reachability, so the robot needs to park as close as possible to the tables without any collisions. An example of this limitation can be seen in Fig.~\ref{fig:nav_manip_challenges} left. For picking the caddy, the arm reaches from above with its wrist in a vertical configuration (Fig.~\ref{fig:nav_manip_challenges} middle). This configuration will cause a navigation-localization challenge and requires the above-mentioned capability. As shown in Fig.~\ref{fig:nav_manip_challenges} right, this capability is also vital for the large gear machining task for two reasons. First, the top door handle on the SCHUNK machine is far and needs to be reachable during the open/close process. Second, the robot needs to see the hole on the lathe chuck for gear insertion. Being able to move the robot closer to the table increases the arm reachability. 
Unlike a conventional navigation problem from A to B in an open space, this task requires frequent stops in front of tables placed at the edge of the map that are sources of navigation challenges. In some cases, the robot might be located very close to two tables. An example can be seen in Fig.~\ref{fig:nav_manip_challenges} left.



\subsection{Adaptation Challenges}

\begin{figure}[t]
\centering
\includegraphics[width=0.9\columnwidth]{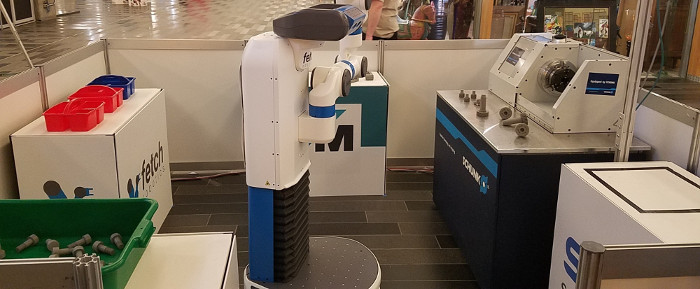} 
\caption{The arena set up during competition. Compared to simulation, the stations are rearranged and a camera pole is placed near the inspection table with the drop-off area marked by black duct tape.}
\label{fig:montreal_arena}
\end{figure}

Even though the regulations are specified in the rule book, there will inevitably be unforeseeable changes that need adaptation. Compared to the simulated environment in Fig. \ref{fig:sim_arena} with the competition setup shown in Fig. \ref{fig:montreal_arena}, the tables are arranged and the walls are added, which did not exist in earlier simulated environments and the Fetch Robotics arena. A camera pole was added near the inspection table with the drop-off area marked by black duct tape. 

For the rearrangement of the tables, we were able to easily change the map poses relative to the wall. To avoid the camera pole, the robot was programmed to proceed and back up by directly commanding the robot base rather than relying on the 2D navigation algorithm.

\section{Discussion and Conclusions}

The navigation, perception and manipulation challenges posed in completing the mobile manipulation tasks in this unique environment are far from solved. Compared to open-space navigation, challenges remain in \textit{autonomous near-obstacle navigation} (e.g., table), which is required for proximity manipulation. For \textit{complex and concave object detection} (e.g., caddy and large gear), future research is needed in benchmarking and developing detection algorithms designed specifically for manipulation. As illustrated in the Fig. \ref{fig:perception_challenges} left, detection becomes more interesting after the object is changed due to manipulation. Finally, \textit{accurate manipulation in a confined operating space} (e.g., large gear insertion) makes obstacle avoidance more challenging.

To promote research in mobile multi-task manipulation, we released our implementation \cite{code} to be used as a baseline. A test course can be easily set up with a few tables and 3D printed parts whose CAD models are available \cite{cad}. The simulation environment is also readily available \cite{fetchitgazebo}.


We have demonstrated our initial efforts towards mobile multi-task manipulation on a kit assembly task with irregular objects and machining in a confined space. We analyzed the challenges that are present in such environments and recommend these as paths for future work.


\section*{Acknowledgements}
This work has been supported in part by the Office of Naval Research (N00014-18-1-2503). Thanks also to the organizers of the FetchIt! Challenge, especially Sarah Elliott, Alex Moriarty, and Melonee Wise.

\addtolength{\textheight}{-4cm}   
\clearpage
\bibliographystyle{IEEEtran}
\bibliography{root}
\end{document}